%% file: main.tex
\PassOptionsToPackage{table}{xcolor}
\documentclass[10pt,twocolumn,letterpaper]{article}

\usepackage[pagenumbers]{iccv} 


\definecolor{blue}{rgb}{0.21,0.49,0.74}
\usepackage[pagebackref,breaklinks,colorlinks,allcolors=blue]{hyperref}
\hypersetup{colorlinks=true, linkcolor=blue, urlcolor=violet}

\usepackage{bbm}
\usepackage{hyperref}
\usepackage{tikz}
\usepackage{comment}
\usepackage{graphicx}
\usepackage{amsmath}
\usepackage{amssymb}
\usepackage{booktabs}
\usepackage{algorithm}
\usepackage{algpseudocode}
\usepackage[accsupp]{axessibility} 
\usepackage{xfrac}
\usepackage{color}
\usepackage{wrapfig}
\usepackage{microtype}
\usepackage{tabularx}
\usepackage{booktabs}
\usepackage{multirow}
\usepackage{blindtext}
\usepackage{xspace}
\usepackage{amsmath}

\definecolor{dimgray}{rgb}{0.41, 0.41, 0.41}

\newcommand{\et}[2]{${#1}^{\pm{#2}}$}

\usepackage[capitalize]{cleveref}
\crefname{section}{Sec.}{Secs.}
\Crefname{section}{Section}{Sections}
\Crefname{table}{Table}{Tables}
\crefname{table}{Tab.}{Tabs.}

\title{SceneMI: Motion In-betweening for Modeling Human-Scene Interactions}

\author{Inwoo Hwang$^{1*}$\quad Bing Zhou$^{2\dagger}$\quad Young Min Kim$^1$\quad Jian Wang$^2$\quad Chuan Guo$^{2\dagger}$ \\ \footnotesize{} \\ \normalsize{$^1$ECE, Seoul National University, $^2$Snap Inc.}}

\begin{document}

\twocolumn[{%
\renewcommand\twocolumn[1][]{#1}%
\maketitle
\begin{center}
    \centering
    \captionsetup{type=figure}
    \includegraphics[width=1.0\linewidth]{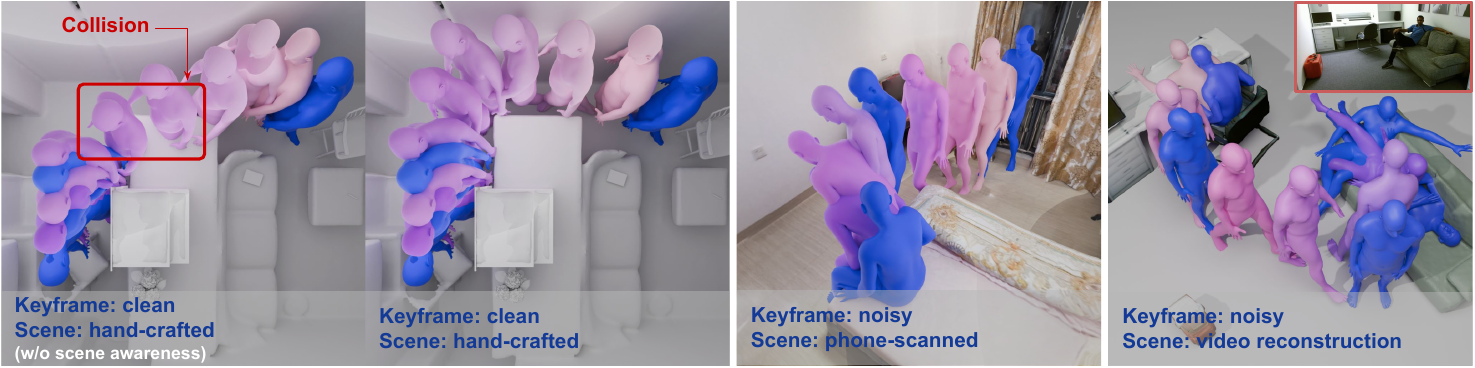}
    \caption{\small \textbf{SceneMI} synthesizes physically plausible transitions (colored in \textcolor{violet}{purple}) that simultaneously satisfy keyframe constraints (colored in \textcolor{blue}{blue}) and environmental affordances in challenging scenarios. The model exhibits robust generalization with \textit{noisy} keyframes in phone-scanned scenes from real-world data, GIMO~\cite{zheng2022gimo} (third figure). SceneMI can further assist realistic 3D human-scene interaction reconstruction only from monocular video inputs (rightmost figure).} 
\label{fig:teaser}
\end{center}
}]

\def\thefootnote{*}\footnotetext{Most work done during an internship at Snap Research NYC, Snap Inc.}
\def\thefootnote{$\dagger$}\footnotetext{Co-corresponding author}

\def\thefootnote{\arabic{footnote}}

\input{sec/0_abstract}
\input{sec/1_intro}
\input{sec/2_related}
\input{sec/3_method}

\input{sec/4_results}
\input{sec/5_conclusion}

{
    \small
    \bibliographystyle{ieeenat_fullname}
    \bibliography{main}
}

\clearpage

\input{sec/X_suppl}

\end{document}

%% file: sec/0_abstract.tex
\begin{abstract}

Modeling human-scene interactions (HSI) is essential for understanding and simulating everyday human behaviors. Recent approaches utilizing generative modeling have made progress in this domain; however, they are limited in controllability and flexibility for real-world applications. To address these challenges, we propose reformulating the HSI modeling problem as Scene-aware Motion In-betweening---a more tractable and practical task. We introduce SceneMI, a framework that supports several practical applications, including keyframe-guided character animation in 3D scenes and enhancing the motion quality of imperfect HSI data. SceneMI employs dual scene descriptors to comprehensively encode global and local scene context. Furthermore, our framework leverages the inherent denoising nature of diffusion models to generalize on noisy keyframes.
Experimental results demonstrate SceneMI's effectiveness in scene-aware keyframe in-betweening and generalization to the real-world GIMO dataset, where motions and scenes are acquired by noisy IMU sensors and smartphones. We further showcase SceneMI's applicability in HSI reconstruction from monocular videos. Project page: {\footnotesize \url{http://inwoohwang.me/SceneMI}}

\end{abstract}

%% file: sec/1_intro.tex
\vspace{-1em}
\section{Introduction}
\label{sec:intro}

Modeling dynamic human motions in everyday environments is challenging due to the inherent complexity of understanding and replicating human-scene interactions (HSI).
Various studies have approached this challenge by learning the data distribution of human motions in scenes through generative modeling based on textual descriptions~\cite{humanise,LINGO,tesmo}, action categories~\cite{TRUMANS,mammos,wang2022diverse}, or past motion sequences~\cite{hassan2021stochastic,MOB,zheng2022gimo,wanglongterm}.
Despite these advances, these methods commonly lack controllability in motions and find limited flexibility in real applications. This motivates our task in this work---scene-aware motion in-betweening, which aims to synthesize natural motion transitions between specific keyframes in 3D scenes while adapting to environmental constraints, such as avoiding obstacles. The availability of keyframe context effectively reduces task complexity while supporting several applications in real scenarios.
For instance, animators can directly incorporate scene information to create 3D character animations from keyframes, as shown in \Cref{fig:teaser}. Additionally, this technique reliably decreases motion artifacts in existing real-world HSI data (\cref{tab:experiments_real}) and enhances HSI reconstruction from monocular videos (\cref{fig:hsi_recon}).

Motion in-betweening has been extensively studied for \textit{isolated} human motions~\cite{harvey2020robust,harvey2018recurrent,zhang2018data,qin2022motion,harvey2020robust,cohan2024flexible}. However, trivially applying these approaches within 3D scenes often leads to undesirable body-scene penetrations.
While some existing scene-aware approaches model goal pose reaching~\cite{wanglongterm,wang2022diverse} as a subsystem component, they typically employ conditional VAEs to capture scene and pose constraints, resulting in limited expressivity and scalability. Another challenge lies in the imperfection of keyframes and scenes in real-world scenarios---such as keyframes captured from noisy motion capture sensors and variations of scenes acquired by different devices.

In this work, we propose SceneMI, a conditional diffusion model tailored for scene-aware motion in-betweening. We extract hierarchical scene features: the global scene is represented by a coarse occupancy voxel grid, while local scenes are encoded as Basis Point Set (BPS) features computed at keyframe time steps based on selected surface body markers. This scene encoding effectively captures the high-level scene layout and spatial context anchored on immediate keyframes, preventing overfitting on specific local scene geometries. We apply imputation to keyframes to effectively encode the temporal location of keyframes within the motion sequence. We particularly leverage the inherent denoising capabilities of diffusion models to handle noisy keyframes. During inference, \textit{noisy} keyframes guide the denoising sampling from steps $T$ until $T^*$ (a hyperparameter). For the remaining steps (from $T^*-1$ to $1$), the full motion sequences, including \textit{noisy} key poses, are iteratively denoised.

In our experiment, SceneMI is trained exclusively on the TRUMANS dataset \cite{jiang2024scaling}, a large-scale HSI dataset with motion capture data and hand-crafted indoor scenes. For evaluation, beyond analysis on TRUMANS, we test the pre-trained model on out-of-domain human-scene interactions from the GIMO dataset \cite{zheng2022gimo}, where motions are captured using inaccurate IMU sensors and scenes are scanned by smartphones in real-world environments.
Empirical results demonstrate that SceneMI generalizes well across different scene types and key pose qualities, excelling at synthesizing smooth transitions. Additionally, SceneMI reliably reduces foot skating by 37.5\% (from 0.261 to 0.163) and jittering by 56.5\% (from 0.573 to 0.249), which are artifacts prevalent in the GIMO dataset. To further explore SceneMI's applicability, we develop a framework to reconstruct human-scene interactions from monocular videos using image-to-3D techniques \cite{xu2024instantmesh,yang2024tencent} and human pose estimation \cite{goel2023humans}. Our results show that SceneMI significantly enhances interaction naturalness and reduces penetration when applied to synthetic 3D scenes and keyframes. 

In summary, this work formally studies the problem of scene-aware motion in-betweening for the first time and explores its application in various HSI scenarios. This is realized by a simple diffusion-based model, SceneMI, which efficiently encodes scene contexts at global and local scales, with specialized denoising procedures to accommodate \textit{noisy} keyframes. Our comprehensive analysis across different scene resources and motion quality, including real-world data, demonstrates SceneMI's motion in-betweening capability and scalability. Additionally, we highlight SceneMI's applicability in reconstructing HSI from monocular video.

%% file: sec/2_related.tex
\section{Related works}
\label{sec:related}

\subsection{Human Motion Synthesis}
Advancements in data-driven approaches have led to appealing breakthroughs in human motion synthesis.
Beyond achieving high-quality and realistic motion outputs, many applications are designed to generate motions adaptable to various conditions, such as action categories~\cite{guo2020action2motion, petrovich2021action, tevet2023human, chen2023executing, hoang2024motionmix}, text~\cite{bae2025moreimprovingmotiondiffusion, chuan2022tm2t, guo2023momask, hwang2025motionsynthesissparseflexible, zhang2023generating, yuan2023physdiff, Guo_2022_CVPR, tevet2023human, zhang2022motiondiffuse, petrovich2022temos, lin2023comes, petrovich24stmc, snapmogen2025}, audio~\cite{lee2019dancing, gong2023tm2d, alexanderson2023listen, siyao2022bailando, tseng2022edge, li2021learn, Qi2023diffdance}, interacting objects~\cite{xu2023interdiff, ghosh2022imos, li2023object, Zhao2022coins, bhatnagar22behave, kulkarni2023nifty, taheri2020grab, bae2023pmp, li2023controllable}, and scenes~\cite{sui2025surveyhumaninteractionmotion, TRUMANS, humanise, LINGO, mammos, lee2023lama, DIMOS, Hwang_2025_CVPR, tesmo, cen2024text_scene_motion, SceneDiffuser, AffordMotion, NEURIPS2024_4620a665, lou2024multimodal}. Recently, there has been growing interest in trajectory-based controllable motion generation. For instance, PriorMDM~\cite{shafir2023human} fine-tuned an existing motion diffusion model MDM~\cite{tevet2023human} to support end-effector trajectory control. GMD~\cite{shafir2023human} incorporated classifier-free guidance into a diffusion model to enable root joint-based controllable generation. OmniControl~\cite{xie2023omnicontrol} expanded this capability to arbitrary joint trajectories using a combination of ControlNet~\cite{zhang2023adding} and inference-time guidance. The recent ControlMM~\cite{pinyoanuntapong2024controlmm} approach, based on masked motion modeling, further enables multi-joint trajectory-based controllable motion generation.

\paragraph{Human-Scene Interaction Modeling.} Despite promising progress in motion synthesis, modeling realistic human movements within environmental scenes remains challenging. Reinforcement learning-based methods~\cite{hassan2021stochastic, zhao2023synthesizing, zhang2022wanderings, SceneDiffuser} have demonstrated the ability to produce goal-oriented locomotion by learning policies with carefully designed reward functions. UniHSI~\cite{xiao2023unified} leverages large language models for detailed motion planning to capture complex human-scene interactions. Another line of research adopts data-driven deep learning models. The complexity of this task often necessitates a hierarchical approach: first generating paths and start/end poses, followed by full motion synthesis~\cite{wang2021scene, wang2021synthesizing, wang2022diverse}.
The latest advances include end-to-end generative models. HUMANISE~\cite{humanise} and LINGO~\cite{LINGO} attempt to generate 3D human motions in scenes from text prompts using diffusion models, while TRUMANS~\cite{TRUMANS} experiments with an autoregressive diffusion model conditioned on action labels. SceneDiffuser~\cite{SceneDiffuser} combines diffusion models with reinforcement learning, extending motion dynamics from a start pose within scenes.
Nevertheless, existing works lack motion controllability and practical applicability, such as recognizing human-scene interactions from real-world demonstrations or aiding in their reconstruction.

\subsection{Motion In-betweening}

Motion in-betweening synthesizes complete motion sequences from keyframes positioned at specific time steps. While traditional approaches use spline interpolation such as Bézier curves~\cite{708559, 10.1145/1073204.1073313, 6909567}, these methods often require extensive post-processing adjustments. Earlier research framed this task as motion planning~\cite{arikan2002interactive, beaudoin2008motion}, employing search and optimization over motion graphs~\cite{kovar2023motion}. However, these approaches typically demand intensive computational resources to maintain and search large databases during inference.
Deep learning has transformed this field. Given the temporal nature of motion data, RNNs have been widely adopted for in-betweening~\cite{harvey2018recurrent, zhang2018data, harvey2020robust}. Transformer architectures have further improved the modeling of long-term dependencies~\cite{duan2021single, qin2022motion, oreshkin2023motion, kim2022conditional}, exemplified by Qin et al.~\cite{qin2022motion}, which developed a two-stage Transformer generating both rough and refined transitions.
While most approaches treat motion in-betweening as deterministic, recent work has explored generative methods. CondMDI~\cite{cohan2024flexible} leverages diffusion models to produce natural, diverse transitions that align with sparse keyframes and optional text instructions. Despite these advancements, a key limitation remains: most existing methods overlook environmental factors that influence motion behavior and assume precise keyframes, which is impractical for real-world inputs.

%% file: sec/3_method.tex
\section{Methodology}

Given a 3D scene $\mathcal{G}$, a sparse set of key poses $\mathbf{s} \in \mathbb{R}^{N \times D}$ with key pose indicator $\mathbf{m}\in\{0, 1\}^N$, our goal is to synthesize a complete motion sequence $\mathbf{x} = \{ x^n \} \in \mathbb{R}^{N \times D}$ that satisfies both the key poses and the environmental constraints of the 3D scene. Here, $N$ represents the total number of poses in the full sequence, with each pose represented by a $D$-dimensional feature vector. 
The mask $m_n \in \mathbf{m}$ indicates whether $n^\text{th}$ frame contains a key pose (${m}_n=1$) or non-key pose (${m}_n=0$), with a total of $k=\sum_n {m}_n \ll N$ key poses.

\paragraph{Motion and Shape Representation} Each pose feature vector comprises global joint positions $J \in \mathbb{R}^{22 \times 3}$ (include root translation $\gamma \in \mathbb{R}^3$), 6D root orientation $\phi \in \mathbb{R}^6$, and local SMPL~\cite{SMPL:2015} pose parameters $\psi \in \mathbb{R}^{21 \times 6}$. These components collectively form a 201-dimensional pose feature representation. To account for body volume variations, we extract human shape features $\mathbf{b}\in \mathbb{R}^{7}$ as distances between a set of representative joint pairs, for example $[\texttt{\small root}, \texttt{\small head}]$, $[\texttt{\small left\_shoulder}, \texttt{\small right\_shoulder}]$.

\begin{figure*}[h]
\centering
\includegraphics[width=1.0\linewidth]{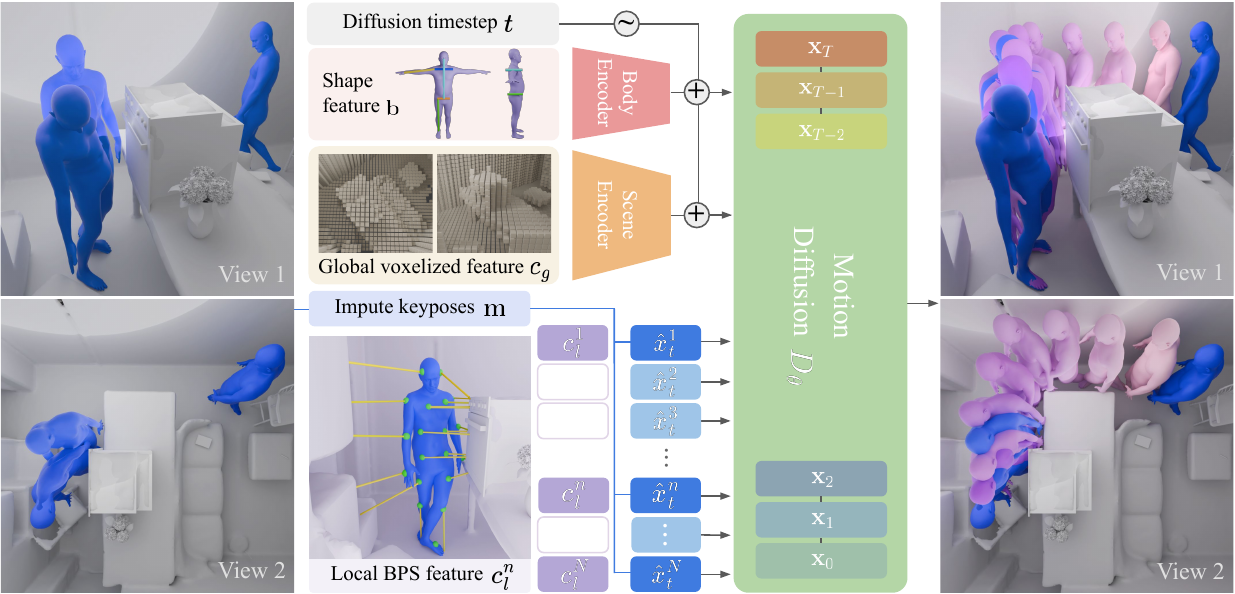}
\caption{\textbf{Overview of SceneMI.} Given the input 3D scene, we extract global voxelized features $\mathbf{c}_g$, and local BPS features $c^n_l$ based on key pose meshes. During training, key poses within the motion sequence are imputed based on the indicator mask $\mathbf{m}$. The model integrates these scene features ($\mathbf{c}_l$ and $\mathbf{c}_g$), body shape features $\mathbf{b}$, and the diffusion timestep $t$ to synthesize the full motion sequence $\mathbf{x}$ that satisfy both the keyframe and scene constraints.}
\label{fig:architecture}
\vspace{-1.2em}
\end{figure*}

\input{sec/3_2_scene_encoding}

\subsection{Scene-aware Motion In-betweening}
\label{sec:SceneMI}

We employ diffusion models for motion in-betweening, which consist of a forward diffuse process and a reverse denoising process. 
Beginning with a conditional data distribution $p(\mathbf{x}_0|\tau)$, where the conditions are the set of scene context $\mathbf{c}$, and body shape $\mathbf{b}$, $\tau=\{\mathbf{c}, \mathbf{b}\}$, and the denoised variable is the clean pose $\mathbf{x}_0=\mathbf{x}$, 
the forward process progressively corrupts original data samples until reaching diffusion step $T$. Noisy samples at timestep $t$ are defined as $\mathbf{x}_t = \sqrt{\bar{\alpha}_t} \mathbf{x}_0 + \sqrt{1 - \bar{\alpha}_t}\boldsymbol{\epsilon}$, where $\bar{\alpha}_t$ is determined by the diffusion noise schedule and $\boldsymbol{\epsilon}$ is sampled from an i.i.d. Gaussian distribution. In the reverse process, the denoising model $\mathcal{D}_{\theta}$ synthesize clean data samples by recursively refining from pure Gaussian noise $q(\mathbf{x}_T)\sim\mathcal{N}(\mathbf{0}, \mathbf{I})$. This process is conditioned on both the timestep $t$ and the corresponding conditional features $\tau$. The denoiser $\mathcal{D}_{\theta}$ is trained to reconstruct the original data $\mathbf{x}_0$:

{\small
\begin{equation}
\label{eq:simple_loss}
\mathcal{L}_\text{simple} = \mathbb{E}_{\mathbf{x}_0 \sim p(\mathbf{x}_0|\tau), t \sim [1, T]} \left[ \left\| \mathbf{x}_0 - \mathcal{D}_\theta(\mathbf{x}_t, t, \tau)) \right\|_2^2 \right]
\end{equation}
}

\paragraph{Model Architecture.} 
Our model employs a U-Net based architecture with Adaptive Group Normalization (AdaGN) and 1D convolution, which has demonstrated strong performance in global motion representation \cite{karunratanakul2023gmd}. AdaGN dynamically adapts normalization, enhancing the model's ability to capture motion dynamics, while 1D convolution facilitates the learning of sequential motion data. The details for the architecture are provided in the supplementary material.

\paragraph{Training Procedure}
During training, we first randomly sample a diffusion timestep $t\sim\mathcal{U}(\{1,..., T\})$. Then, $k\sim\mathcal{U}(\{2,...,N\})$ keyframes are randomly selected from the motion sequence, including the start and end frames, forming a binary mask $\mathbf{m}\in\{0, 1\}^N$ where ${m}_n=1$ indicates a keyframe location. Using this mask $\mathbf{m}$, we perform imputation on the noisy sample $\mathbf{x}_t$ by substituting the feature values at keyframe locations with their corresponding clean values from $\mathbf{x}_0$
, as follows:
\[
\mathbf{x}_t' = \mathbf{m} \odot \mathbf{x}_0 + (1 - \mathbf{m}) \odot \mathbf{x}_t
\]
where $\odot$ denotes element-wise multiplication.
And we retain local scene features $\mathbf{c}_l$, which are visible from selected keyframes:
\[
\mathbf{c}_l' = \mathbf{m} \odot \mathbf{c}_l 
\]

The motion-related features are obtained by concatenating the features of $\tilde{\mathbf{x}}_t = \texttt{\small spatialconcat}(\mathbf{x}_t', \mathbf{c}_l', \mathbf{m})$. These features are further concatenated with body shape features $\mathbf{b}$ and global scene features $\mathbf{c}_g$ along the temporal dimension. Subsequently, the embeddings of diffusion steps (i.e., $t$) are added to all input features, as illustrated in~\cref{fig:architecture}. During training, we randomly mask out $\mathbf{c}_g$ with a probability of $10\%$ to enable classifier-free guidance during inference.

In addition to the reconstruction loss $\mathcal{L}_{\text{simple}}$  defined in Eq.~\ref{eq:simple_loss}, our denoiser $D_\theta$ is trained with additional losses on the global joint positions and joint velocities to enhance realism, which are obtained through forward kinematics (FK) of the predicted SMPL parameters, as following:

\vspace{-15pt}
{\small
\begin{align}
\mathcal{L}_{\text{joints}} &= \left\| \texttt{FK}(\mathbf{x}_0) - \texttt{FK}(\mathcal{D}_\theta(\mathbf{x}_t, t, \tau)) \right\|^2, \\
\mathcal{L}_{\text{vel}} &= \left\| \texttt{diff}(\texttt{FK}(\mathbf{x}_0)) - \texttt{diff}(\texttt{FK}(\mathcal{D}_\theta(\mathbf{x}_t, t, \tau))) \right\|^2.
\end{align}
}

Overall, the final learning objective is:
\begin{equation}
\mathcal{L} = \mathcal{L}_{\text{simple}} + \lambda_{\text{joints}} \mathcal{L}_{\text{joints}} + \lambda_{\text{vel}} \mathcal{L}_{\text{vel}}.
\end{equation}

\paragraph{Inference}
At inference, the goal is to synthesize the dense motion sequence $\mathbf{x}_0$ from the keyframes $\mathbf{s}$, along with the given keyframe indices $\mathbf{m}$, where $\mathbf{s}_n=\mathbf{x}_0^n$ if $\mathbf{m}_n=1$. We directly calculate the global feature $\mathbf{c}_g$ and the keyframe-centered local features $\mathbf{c}'_l$. At each sampling step, we impute the keyframe features following the same procedure used during training.
\[
\mathbf{x}_t' = \mathbf{m} \odot \mathbf{s} + (1 - \mathbf{m}) \odot \mathbf{x}_t.
\]

Then we concatenate all motion-related features $\tilde{\mathbf{x}}_t = \texttt{\small spatialconcat}(\mathbf{x}_t', \mathbf{c}'_l, \mathbf{m})$ and apply classifier-free guidance with scale $\mathbf{w}$ to the denoiser output:
\[
\hat{\mathbf{x}}_0 = \mathbf{w} \cdot \mathcal{D}_\theta(\tilde{\mathbf{x}}_t, t, \mathbf{b}, \mathbf{c}_g) + (1 - \mathbf{w}) \cdot \mathcal{D}_\theta(\tilde{\mathbf{x}}_t, t, \mathbf{b}, \emptyset)
\]

\subsubsection{Handling Noisy Keyframes}
\label{sec:Learning_Noisy}

In real-world scenarios, when the keyframe poses are captured by less accurate mocap sensors or extracted from videos, directly using \textit{noisy} keyposes for motion inbetweening would compromise the output motion quality. Fortunately, diffusion models inherently add noise to motion data during the forward process, and the models learn to remove this noise during the reverse process. As suggested by MotionMix~\cite{hoang2024motionmix} and Rohm~\cite{zhang2024rohm}, the noise in imperfect motions could be analogous to the added noise in the diffusion process at certain timesteps. Motivated by this insight, we divide the diffusion/sampling timesteps into two ranges: $[T, T^*+1]$ and $[T^*, 1]$. From early diffusion timestep $T$ to $T^*+1$, the prediction of $\mathbf{x}_0$ is guided by noisy keyposes $\mathbf{s}^\text{noisy}$ for keyframe alignment, while for the remaining timesteps ($T^*$ to $1$), both keyframes and in-betweening frames are iteratively denoised by $\mathcal{D}_{\theta}$.

To learn motion inbetweening with noisy keyframes, we first create a noisy motion dataset $\mathbf{x}_0^{\text{noisy}}$ from clean motions $\mathbf{x}_0$ by adding unitary Gaussian noise scaled by a random factor $l \sim \mathcal{U}(0, 1.0)$. During training, we restrict the imputation operation to only occur until timestep $T^*$:
{\small
\[
\mathbf{x}_t' =
\begin{cases}
\mathbf{m} \odot \mathbf{x}_0^{\text{noisy}} + (1 - \mathbf{m}) \odot \mathbf{x}_t, & t \in [T, T^* + 1] \\
\mathbf{x}_t, & t\in [T^*, 1] \\
\end{cases}
\]
}
The training objective is to predict the clean motion $\mathbf{x}_0$.

During inference, we apply the same imputation strategy to noisy keyframes $\mathbf{s}^{noisy}$ for sampling:
{\small
\[
\mathbf{x}_t' =
\begin{cases}
\mathbf{m} \odot \mathbf{s}^{\mathrm{noisy}} + (1 - \mathbf{m}) \odot \mathbf{x}_t, & t \in [T, T^* + 1] \\
\mathbf{x}_t, & t\in [T^*, 1] \\
\end{cases}
\]
}

Empirically, we experiment with different $T^*$ values (see \Cref{tab:experiments_noise}) and find $T^*=20$ performs the best, where the total diffusion step $T=1000$.

\input{tab/mib_scene_merged}

%% file: sec/3_2_scene_encoding.tex
\subsection{Scene Encoding}
\label{sec:scene_encoding}

We need to provide a compact yet descriptive scene representation as the condition of our motion in-betweening module.
As shown in \cref{fig:architecture} (middle block), we adopt distinct strategies to encode scenes at global and local scales. The global scene feature $\mathbf{c}_g$ encapsulates the expansive scene layout, providing crucial information for navigating the overall motion trajectory. The local scene features $\mathbf{c}_l$ encode spatial information centered around individual keyframes, enabling fine-grained interactions with the environment. Together, these features $\mathbf{c} = [\mathbf{c}_g, \mathbf{c}_l]$ provide a comprehensive environmental context that guides motion in-betweening, which also presents strong generalization capability to out-of-domain 3D scenes.

\vspace{-1em}
\paragraph{Global Scene Features}

The global scene features represent the overall spatial context of the entire scene as an occupancy voxel grid at a coarse resolution of $\mathbf{c}_g \in \{0, 1\}^{d_x \times d_y \times d_z}$ with 0.1m per voxel, where a value of 1 indicates an occupied voxel and 0 represents an unoccupied space. 
The center and orientation of this grid are initialized by the root position and orientation in the first frame. 
To effectively encode these global scene features, we employ a Vision Transformer (ViT) model, following the approach in \cite{TRUMANS}. This global scene encoder processes inputs with dimensions $48\times24\times48$ and results in a 512-dimensional feature vector. The global scene encoder is jointly trained with the rest of the framework.

\paragraph{Keyframe-centered Local Scene Features}
We extract local scene features using the Basis Point Set (BPS)~\cite{bps} approach surrounding the keyframe poses. First, we determine 64 anchor points on the T-posed SMPL mesh surface via farthest-point sampling, creating a structured abstraction of the body shape geometry. These anchor vertex indices remain fixed across all keyframe poses throughout our experiments. For each keyframe, we calculate the nearest point from the 3D scene to each indexed anchor point on the body mesh. The resulting ordered offset vectors—relative positions to these nearby points—serve as the BPS features $\mathbf{c}^n_l\in\mathbb{R}^{64\times 3}$ for the immediate local scene surrounding each keyframe. These features are embedded through an MLP and then concatenated with key pose features at their corresponding frames. Compared to traditional geometry representation, BPS features specifically target scene context close to the human body while remaining agnostic to point order, mesh topology, and resolution, thereby enhancing generalizability on different scene sources.

%% file: tab/mib_scene_merged.tex
\begin{table*}[t!]
    \centering
\resizebox{0.92\linewidth}{!}{
\begin{tabular}{lccccccc}
\toprule
\multirow{2}{*}{\centering Method} & \multirow{2}{*}{\centering FID~$\downarrow$} & \multicolumn{1}{p{1.5cm}}{\centering Foot Skating $\downarrow$} & \multicolumn{1}{p{1.5cm}}{\centering Jerk ($m/s^3$) $\downarrow$} & \multicolumn{1}{p{1.5cm}}{\centering MJPE Key ($m$) $\downarrow$}  & \multicolumn{1}{p{1.6cm}}{\centering MJPE All ($m$) $\downarrow$} & \multicolumn{1}{p{2.3cm}}{\centering Collision Frame Ratio  $\downarrow$} & \multicolumn{1}{p{1.8cm}}{\centering Pene Max ($m$) $\downarrow$} \\
\midrule
MDM~\cite{tevet2023human}                      & 1.422 & 0.316 & 0.972 & 0.568 & 0.576 & 0.317  & 0.112   \\
StableMoFusion~\cite{huang2024stablemofusion}  & 0.732 & 0.264 & 0.272 & 0.412 & 0.471 & 0.275  & 0.098   \\
SceneDiffuser~\cite{SceneDiffuser}             & 1.397 & 0.432 & 0.583  & 0.349 & 0.391  & 0.292  & 0.128   \\
Wang et al.~\cite{wang2020synthesizing}        & 3.243 & 0.528  & 17.243 & 0.091 & 0.096 & 0.203  & 0.082   \\
OmniControl~\cite{xie2023omnicontrol}          & 0.371 & 0.294 & 0.274 & 0.217 & 0.294 & 0.211 & 0.081   \\
CondMDI~\cite{cohan2024flexible}               & 0.943 & 0.281 & 0.305 & 0.452 & 0.457 & 0.262 & 0.087   \\
\midrule
Ours                                           & \textbf{0.123} & \underline{0.248} & 0.194 & \textbf{0.006} & \textbf{0.023} & \textbf{0.113} & \textbf{0.043}   \\
\;\; w/o scene-awareness $\mathbf{c}_g, \mathbf{c}_l$   & 0.136     & 0.251 & \textbf{0.103}& 0.012 & 0.059 & 0.131 & 0.049  \\
\;\; w/o global feature $\mathbf{c}_g$        &  0.138           & 0.254 & \underline{0.131}  & 0.011 & 0.051  & 0.128 & 0.048   \\
\;\; w/o local feature $\mathbf{c}_l$          & \underline{0.125}  & \textbf{0.245}  & 0.196  & \underline{0.008} & \underline{0.036} & \underline{0.119} & \underline{0.045}  \\

\bottomrule
\end{tabular}
}
\caption{Quantitative scene-aware motion in-betweening results on TRUMANS dataset~\cite{TRUMANS} with \textcolor{red}{noise-free} keyframes. Our method excels in in-betweening within scene constraints across various metrics. The keyframe interval is set to $r=60$ frames. \textbf{Bold} represents the best value, and \underline{underlined} represents the second-best.}
\label{tab:experiments_scene}
\vspace{-0.5em}
\end{table*}

%% file: sec/4_results.tex
\begin{figure*}[h]
\centering
\includegraphics[width=1.0\linewidth]{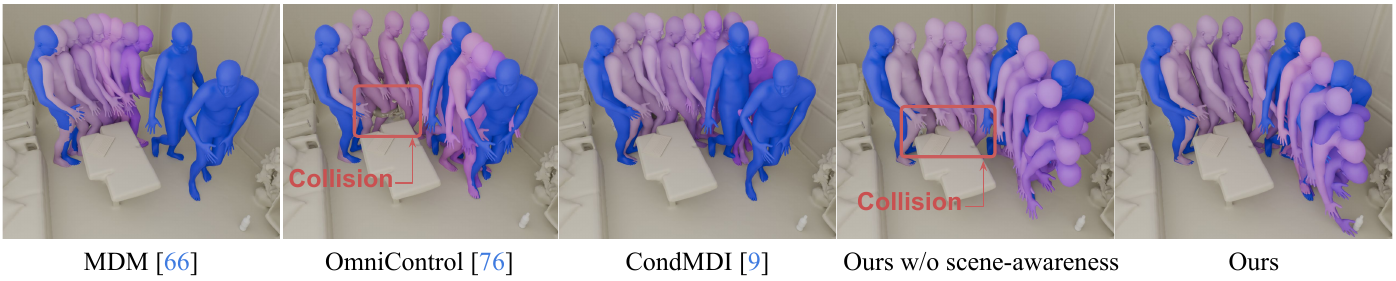}
\caption{\textbf{Visual comparisons} with baseline models on \textcolor{red}{noise-free} TRUMANS dataset. }
\label{fig:trumans_results}
\vspace{-0.7em}
\end{figure*}

\section{Experiments}
We provide a comprehensive evaluation of our approach in three settings: motion in-betweening with noise-free keyframes and in-domain 3D scenes in~\cref{sec:exp_scene}, out-of-domain noisy keyframes and real-world scenes in~\cref{sec:exp_noisy}, and applications for refining noisy human-scene interactions in~\cref{sec:exp_hsi_rec}.

\paragraph{Dataset.} For training, we utilized TRUMANS~\cite{TRUMANS}, the largest high-quality motion capture dataset (30 FPS) with hand-crafted scene geometry. Motion sequences were segmented into $N=121$ frames, with 10\% randomly selected for the test set. To train for noisy keyframe handling, we added zero-mean noise with a standard deviation determined by noise level $l$, defined as $(l^{\circ}, l^{\circ}, l \ \text{cm})$ for the ground-truth SMPL parameters$(\phi, \psi, \gamma)$ except the start and end frames.
For evaluation, we tested our approach on both noise-free and synthetic noisy TRUMANS test sets. To further assess out-of-domain generalization robustness, we evaluated performance on the real-world GIMO dataset~\cite{zheng2022gimo} and demonstrated video-based HSI reconstruction using the PROX dataset~\cite{PROX}.

\paragraph{Implementation Details.}
We implement our pipeline using PyTorch~\cite{pytorch} and train our model on a single NVIDIA RTX 3090 GPU. During training, we use $T=1000$ diffusion timesteps and set $\lambda_{\text{joints}} = 2.0$ and $\lambda_{\text{vel}} = 10.0$, with 
$T^*=20$ for the noisy setting. During inference, we use a guidance scale of $\mathbf{w}=2.5$ with $T=1000$ diffusion sampling steps. 

\paragraph{Evaluation Metrics.} 

We evaluate motion in-betweening results using three sets of metrics that assess motion naturalness, motion alignment, and human-scene interaction plausibility. For motion naturalness, we calculate the \textit{Frechet Inception Distance (FID)}, which measures the distributional distance between high-level features of completed motions and ground truth data. We also quantify \textit{foot skating} and \textit{jerk} (the rate of change of acceleration) as indicators of motion naturalness. For motion alignment, we compute the \textit{mean joint position error} at keyframe locations (MPJE Key) and across the full sequence (MPJE All). To evaluate HSI plausibility, we report the \textit{collision frame ratio}---proportion of frames containing human-scene collisions relative to the total sequence length, and the \textit{penetration max}---average maximum collision distance across each interaction sequence.

\paragraph{Baselines.}

Since scene-aware motion in-betweening has not been formally investigated in previous works, we adapt several state-of-the-art approaches from related fields as our baselines. These include diffusion models for scene-agnostic motion synthesis (MDM~\cite{tevet2023human} and StableMoFusion~\cite{huang2024stablemofusion}), motion in-betweening methods (OmniControl~\cite{xie2023omnicontrol} and CondMDI~\cite{cohan2024flexible}), and scene-aware motion synthesis approaches (SceneDiffuser~\cite{SceneDiffuser} and Wang et al.~\cite{wang2020synthesizing}). Baselines that originally did not incorporate scene information~\cite{tevet2023human, huang2024stablemofusion, xie2023omnicontrol, cohan2024flexible} were re-trained with additional \textit{global scene conditions}. Detailed descriptions of all baseline implementations are provided in the supplementary material.

\subsection{Evaluation on Noise-free TRUMANS}
\label{sec:exp_scene}

We first evaluate SceneMI on the TRUMANS~\cite{TRUMANS} dataset using the classical motion in-betweening setting with surrounding environments, where keyframes are clean, and 3D scenes belong to the same domain as the training data. In our experiments, we employ a sparse keyframe interval of 60 frames, corresponding to 2-second motion sequences between consecutive keyframes.

\paragraph{Comparison with Baselines.}

Table~\ref{tab:experiments_scene} presents the quantitative results. Our model demonstrates superior performance across all evaluation metrics, achieving the highest motion quality (FID of 0.123 compared to CondMDI), precise keyframe pose alignment (MJPE of 0.023 mm
m), and enhanced environmental awareness as evidenced by the lowest collision frame ratio (0.113). In contrast, general motion synthesis models such as MDM~\cite{tevet2023human}, StableMoFusion~\cite{huang2024stablemofusion}, and SceneDiffuser~\cite{SceneDiffuser} lack effective in-betweening capabilities, particularly in keyframe alignment, as they primarily focus on modeling motion distribution rather than in-betweening functionality. Wang et al.
\cite{wang2020synthesizing} produce lower motion quality (FID of 3.243) due to their reliance on CVAE and global point cloud features, which have limited expressivity. OmniControl\cite{xie2023omnicontrol} struggles to effectively incorporate scene information, resulting in observable body-scene collisions, as shown in~\Cref{fig:trumans_results}. Similarly, CondMDI~\cite{cohan2024flexible} exhibits inferior performance in both motion quality (FID of 0.943) and keyframe alignment (see~\cref{fig:trumans_results}), attributable to its dependence on motion velocity features as model inputs, which are rarely available in practical applications.

\input{tab/mib_challenge_frames}

\input{tab/mib_noise}

\paragraph{Ablation Analysis.} We compared our model to ablated variants that exclude global scene features $\mathbf{c}_g$, local scene features $\mathbf{c}_l$, or both (i.e.,``w/o scene awareness``). We also evaluated performance on close-proximity interaction frames where the minimal distance between human and scene is less than 8 cm, with results shown in \cref{tab:experiments_challenging_frames}. From \cref{tab:experiments_scene} and \cref{tab:experiments_challenging_frames}, we observed that scene-awareness effectively reduces the collision ratio from 0.237 to 0.162 in interaction frames and enhances alignment with ground-truth motions by over 50\% (MJPE from 0.059 to 0.023 m). This observation is consistent with the visualized generation results in \cref{fig:trumans_results}. Both global and local scene features enhance motion quality and alignment, with global features playing more prominent roles. Refer to the supplementary video for more results.

\input{tab/mib_real_world}

\subsection{Robust Generalization to Noisy Data}
\label{sec:exp_noisy}

Next, we demonstrated SceneMI's ability to generalize from synthetic to real-world noisy data. To enhance our model's robustness to noise, we trained exclusively on TRUMANS by introducing synthetic noise and applying a diffusion process divided into two ranges:  $[T, T^*+1]$ and $[T^*, 1]$, with $T^*=20$, as described in \Cref{sec:Learning_Noisy}.

We first evaluated SceneMI's performance on the noisy TRUMANS test set, where synthetic noise was added to keyframes. Then, we extended our evaluation to the real-world GIMO dataset~\cite{zheng2022gimo}, which contains natural noise characteristics arising from inaccurate motion capture sensors. Additionally, GIMO's scenes are scanned by smartphone, which significantly differs from the synthetic scenes in TRUMANS that our model was trained on.

\paragraph{Evaluation on TRUMANS with Synthetic Noise.}
We employed a dense keyframe interval of 3 frames, with a maximum noise level $l=1$ for evaluation. Dense, noisy keyframes were intentionally used for motion in-betweening, which typically presents a more challenging scenario than in-betweening with sparse, noisy keyframes. Table~\ref{tab:experiments_noise} presents the quantitative results. Our model consistently outperforms baseline models across various metrics. Baseline models that directly use noisy keyframes for motion in-betweening face degraded motion quality (e.g., MDM with an FID of 5.149), even with denser keyframes compared to the noise-free setting in Table~\ref{tab:experiments_scene}. In comparison, our noise-aware approach steadily produces high-quality motions (e.g., FID of 0.118) while respecting both keyframes (MJPE of 0.012 m) and environmental constraints.

In addition, we validate our noise-aware design in Section~\ref{sec:Learning_Noisy}, and experiment with different $T^*$ values, as shown in Table~\ref{tab:experiments_noise}. Noise awareness ($T^*\neq 0$) consistently improves motion in-betweening performance on noisy keyframes. When setting $T^*=20$, our approach produces motions with FID improved from 0.157 to 0.118, and jerk reduced from 0.230 to 0.198. Generally, a larger $T^*$ yields smoother motion with reduced jerk; however, it also increases keyframe error due to the randomness introduced by the diffusion sampling process in the range  $[T^*, 1]$. We find $T^*=20$ as a sweet spot value, synthesizing high-quality motion while preserving keyframe accuracy.

\begin{figure}[tb]
\centering
\includegraphics[width=0.99\linewidth]{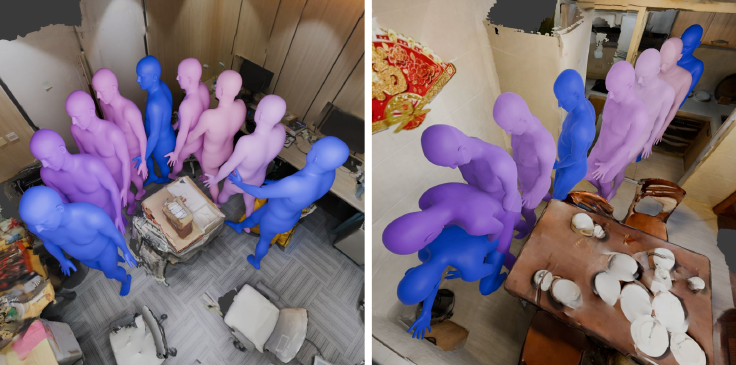}
\vspace{-0.5em}
\caption{SceneMI can perform motion in-betweening on the real-world GIMO dataset~\cite{zheng2022gimo}, while maintaining the original semantics in a scene-aware manner.}
\label{fig:real_world_results}
\vspace{-1.5em}
\end{figure}

\paragraph{Generalization on Real-World HSI.}

To emphasize the robust and generalizable performance of SceneMI under noisy conditions, we evaluate our algorithm on the real-world GIMO dataset. \Cref{fig:real_world_results} presents qualitative visualizations, showing SceneMI generating in-betweening walking motions around indoor tables and chairs. Table~\ref{tab:experiments_real} reports quantitative results for both baseline models and our approach, including ablated versions. We omit keyframe \textit{MJPE} and \textit{FID} scores due to the unavailability of ground-truth motion. Though exclusively trained on the TRUMANS dataset, SceneMI demonstrates strong generalizability to the noisy real-world keyframes in GIMO, outperforming baselines by a large margin. Compared to the original GIMO dataset, the in-betweening motions from SceneMI exhibit significantly reduced \textit{foot skating} (from 0.261 to 0.163) and \textit{jittering} (from 0.573 to 0.249), demonstrating SceneMI's capability to improve imperfect HSI data.

Ablation analysis provides further insights into how scene-awareness and noise-awareness contribute to performance. Our results confirm our design motivation: \textit{noise-awareness} primarily enhances motion quality, while \textit{scene-awareness} effectively reduces the incidence of collision.

\subsection{Application: Video-based HSI Reconstruction}
\label{sec:exp_hsi_rec}

We highlight SceneMI's applicability and generalizability by applying it to reconstructed geometry and estimated keyposes from real-world monocular videos. Given one frame from a monocular video, we first segment each instance~\cite{cheng2021mask2former} and use off-the-shelf image-to-3D techniques~\cite{xu2024instantmesh,yang2024tencent} to reconstruct individual objects. This is followed by estimating depths~\cite{depthanything} and camera parameters~\cite{jin2023perspective} for coordinate calibration to properly place each object according to the layout in the video frame. We then recover the human mesh from the video stream~\cite{goel2023humans} and place the motion sequence in the reconstructed scene according to camera parameters.

\begin{figure}[t!]
\centering
\includegraphics[width=1.0\linewidth]{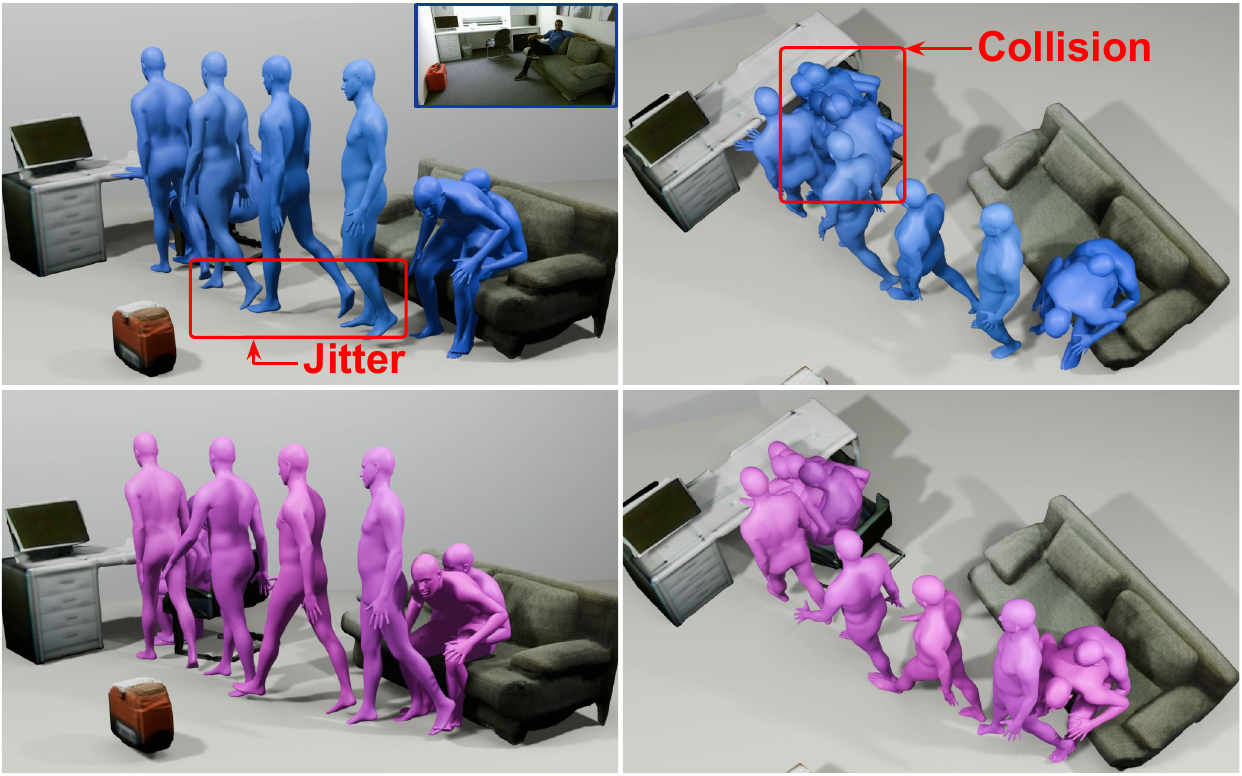}
\vspace{-1.7em}
\caption{SceneMI can be applied to reconstructed scenes and keyframes from video, facilitating realistic and physically plausible human-scene interaction reconstruction from monocular video.} 
\label{fig:hsi_recon}
\vspace{-1.5em}
\end{figure}

Initially, only a partial scene can be reconstructed. Motions also contain severe jitters and foot skating due to occlusion and depth ambiguity. Moreover, since scene geometry and motions are obtained independently, they exhibit serious collisions, as shown in \cref{fig:hsi_recon} (top). To refine the results, we uniformly select keyframes with 15-frame intervals and apply SceneMI to re-generate the motion sequence within synthetic scenes through keyframe in-betweening. The post-processed motions are much smoother, containing fewer motion artifacts and maintaining more plausible interactions with scenes, as shown in \cref{fig:hsi_recon} (bottom). Additionally, we employ an autoregressive sampling strategy to synthesize \textit{long-duration} sequences. By iteratively using the final $v=60$ frames of a prior episode as initial keyframes for the subsequent segment, we synthesize continuous motion sequences across a 23-second video (30 FPS, 690 frames), as demonstrated in the supplemental videos.

To our knowledge, we are the \text{first} to build a complete pipeline for 3D HSI reconstruction from monocular videos. Since this is not the focus of the current work, we elaborate the detailed procedure in the appendix.

%% file: tab/mib_challenge_frames.tex
\begin{table}[t!]
    \centering
\resizebox{0.99\linewidth}{!}{
\begin{tabular}{lccc}
\toprule
\multirow{2}{*}{Method} & \multicolumn{1}{p{2.3cm}}{\centering Collision Frame Ratio  $\downarrow$} & \multicolumn{1}{p{2.3cm}}{\centering Collision Vertex Ratio~$\downarrow$} & \multicolumn{1}{p{1.5cm}}{\centering Pene. Max ($m$)~$\downarrow$} \\
\midrule
Ours w/o scene-aware & 0.237 & 0.035 & 0.056   \\
Ours & \textbf{0.162} & \textbf{0.020} & \textbf{0.047}  \\
 \bottomrule
\end{tabular}
}
\caption{Quantitative evaluation on the \textit{close-proximity} human-scene interaction frames from the TRUMANS~\cite{TRUMANS}.}
\label{tab:experiments_challenging_frames}
\vspace{-1.2em}
\end{table}

%% file: tab/mib_noise.tex
\begin{table*}[ht!]
    \centering
\resizebox{0.92\linewidth}{!}{
\begin{tabular}{lccccccc}
\toprule
\multirow{2}{*}{\centering Method} & \multirow{2}{*}{\centering FID~$\downarrow$} & \multicolumn{1}{p{1.5cm}}{\centering Foot Skating $\downarrow$} & \multicolumn{1}{p{1.5cm}}{\centering Jerk ($m/s^3$) $\downarrow$} & \multicolumn{1}{p{1.5cm}}{\centering MJPE Key ($m$) $\downarrow$}  & \multicolumn{1}{p{1.6cm}}{\centering MJPE All ($m$) $\downarrow$} & \multicolumn{1}{p{2.3cm}}{\centering Collision Frame Ratio  $\downarrow$} & \multicolumn{1}{p{1.8cm}}{\centering Pene. Max ($m$) $\downarrow$} \\
\midrule
MDM~\cite{tevet2023human} & 5.149 & 0.761 & 22.169 & 0.285 & 0.279 & 0.151 & 0.056   \\
OmniControl~\cite{xie2023omnicontrol} & 2.981 & 0.381 & 2.198 & 0.302 & 0.308 & 0.169  &  0.058 \\
CondMDI~\cite{cohan2024flexible} & 3.136 & 0.317 & 0.296 & 0.354 & 0.349 & 0.187 & 0.059  \\
\midrule
Ours ($T^*=20$)                  & \textbf{0.118} & \textbf{0.247} & 0.198            & \textbf{0.013} & \textbf{0.012} & \textbf{0.108}  & \textbf{0.042}   \\
\;\; $T^*=0$ (w/o noise-awareness)   & 0.157          & 0.265          & 0.230            & 0.015          & 0.014          & 0.119           & 0.046  \\
\;\; $T^*=10$                    & 0.123        & 0.253            & 0.199            & \underline{0.013}          & \underline{0.013}   &  \underline{0.110}        & \underline{0.044}  \\
\;\; $T^*=40$                    & \underline{0.121}  & \underline{0.249}            & \textbf{0.187}   & 0.014          & 0.014          & 0.112        & 0.045  \\
\;\; $T^*=60$                    & 0.122        & 0.250            & \underline{0.189}            & 0.015          & 0.015          & 0.114        & 0.045  \\

 \bottomrule
\end{tabular}
}
\vspace{-0.5em}
\caption{Quantitative scene-aware motion in-betweening results TRUMANS dataset~\cite{TRUMANS} with \textcolor{blue}{synthetic noise}. keyframes are provided, using an interval of $r = 3$ and a noise level of $l = 1$. We intentionally used dense keyframes, which presents a more challenging scenario than dealing with sparse, noisy keyframes. \textbf{Bold} represents the best value, and \underline{underlined} represents the second-best.
}
\label{tab:experiments_noise}
\vspace{-1.3em}
\end{table*}

%% file: tab/mib_real_world.tex
\begin{table*}[ht!]
    \centering
\resizebox{0.99\linewidth}{!}{
\begin{tabular}{lccccc}
\toprule
\multicolumn{1}{p{2.6cm}}{ Method} &  \multicolumn{1}{p{2.3cm}}{\centering Foot Skating $\downarrow$} & \multicolumn{1}{p{2.3cm}}{\centering Accel ($m/s^2$) $\downarrow$} & \multicolumn{1}{p{2.3cm}}{\centering Jerk ($m/s^3$) $\downarrow$} & \multicolumn{1}{p{3.4cm}}{\centering Collision Frame Ratio $\downarrow$} & \multicolumn{1}{p{3.3cm}}{\centering Pene Max ($m$)~$\downarrow$} \\
\midrule
\rowcolor{orange!30}
Real-World Data (GIMO~\cite{zheng2022gimo}) & 0.261 & 0.347 & 0.573   & 0.057  &  0.048 \\
\midrule
MDM~\cite{tevet2023human} & 0.329 & 0.491  & 0.828 &   0.132   & 0.114   \\
OmniControl~\cite{xie2023omnicontrol} & 0.301 & 0.413 & 0.624 &  0.102 &  0.092  \\
CondMDI~\cite{cohan2024flexible} & 0.312 & 0.359 &  0.498  &  0.091 & 0.083   \\
\midrule
Ours (full) & \textbf{0.163} & \underline{0.165} &  \underline{0.249} & \textbf{0.060}   &   \textbf{0.047} \\
\;\; w/o scene awareness & 0.192 & \textbf{0.163} & \textbf{0.245} & 0.082 & 0.051 \\
\;\;  w/o noise awareness & 0.391 & 0.215 & 0.301 & 0.072 & 0.049 \\

 \bottomrule
\end{tabular}
}
\vspace{-0.5em}
\caption{Quantitative evaluation on \textcolor{orange}{real-world} GIMO~\cite{zheng2022gimo}, which naturally contains noise arising from acquisition equipment, using an interval of $r=15$. Through motion in-betweening, our method demonstrates the ability to reduce \textit{foot skating} and \textit{jerk} that are prevalent in the original motion data. \textit{Noise awareness} plays a key role in improving motion quality while \textit{scene-awareness} effectively reduces collisions. \textbf{Bold} represents the best value, and \underline{underlined} represents the second-best.
}
\label{tab:experiments_real}
\vspace{-1.2em}
\end{table*}

%% file: sec/5_conclusion.tex
\section{Conclusions}
We propose SceneMI, a scene-aware motion in-betweening framework for modeling human-scene interactions (HSI). SceneMI  synthesizes smooth and natural human motions by processing both clean and noisy keyframes within surrounding geometric contexts. This capability enables practical applications including keyframe-guided character animation in 3D scenes and enhances motion quality from imperfect HSI data captured via noisy sensors or video reconstructions. Our experiments demonstrate that SceneMI not only excels in classical in-betweening scenarios with scene constraints but also robustly generalizes to real-world GIMO data. Additionally, we showcase SceneMI's versatility through a new HSI reconstruction pipeline for monocular video.

\paragraph{Limitations.} SceneMI can be improved in several aspects: (i) it relies on full-pose keyframes, which limits flexibility when only partial poses are available; (ii) scene-aware motion in-betweening conditioned on text input could enhance controllability; and (iii) SceneMI models human-scene interactions primarily through feature-level fusion, potentially restricting the expressivity. Future work should consider model-level fusion modeling to address these limitations.

\paragraph{Acknowledgements}
We sincerely thank the Computational Imaging team at Snap Research NYC, Snap Inc., for their generous support and valuable collaboration.
This work was partly supported by IITP grant funded by the Korea government (MSIT) [NO.RS-2021-II211343, Artificial Intelligence Graduate School Program (Seoul National University)] and BK21 FOUR program of the Education and Research Program for Future ICT Pioneers, Seoul National University in 2025.

%% file: sec/X_suppl.tex
\maketitlesupplementary

In the supplementary materials, we elaborate the implementation details for our SceneMI (Sec.~\ref{sec:suppl_details}), additional analysis with experiments on varying keyframe selection strategy, runtime analysis, and an ablation study on hyperparameter settings with discussing limitations (Sec.~\ref{sec:add_details}). Furthermore, we introduce a detailed \textit{Video-based Human-Scene Interaction Reconstruction} pipeline (Sec.~\ref{sec:suppl_video2animation}), where SceneMI plays a crucial role in enhancing realism and physical plausibility in HSI reconstruction. For additional qualitative results, please refer to the supplementary video on our project page.

\section{Further Details}
\label{sec:suppl_details}

\subsection{Implementation Details}

We implemented our model using a DDPM based diffusion framework~\cite{ho2020denoisingdiffusionprobabilisticmodels}, leveraging the U-Net architecture proposed by \cite{karunratanakul2023gmd} with the AdamW optimizer~\cite{loshchilov2019decoupledweightdecayregularization} with a learning rate of \( 1\mathrm{e}{-4} \) and a weight decay of \( 1\mathrm{e}{-2} \). For classifier-free guidance at inference, we set the guidance weight \( \mathbf{w} = 2.5 \).
More hyperparameters of the architecture and diffusion process are organized in Table~\ref{table:hyperparameters}.

\input{tab/hyperparameter}

\subsection{Baseline Details}

We compare our approach against a diverse range of state-of-the-art motion synthesis methods, including scene-agnostic motion generation (MDM~\cite{tevet2023human} and StableMoFusion~\cite{huang2024stablemofusion}), motion in-betweening (OmniControl~\cite{xie2023omnicontrol} and CondMDI~\cite{cohan2024flexible}), and scene-aware motion synthesis (SceneDiffuser~\cite{SceneDiffuser} and Wang et al.~\cite{wang2020synthesizing}). 
To ensure a fair comparison of scene-aware motion in-betweening tasks, we adapt their original models accordingly.

For scene agnostic works (MDM~\cite{tevet2023human}, StableMoFusion~\cite{huang2024stablemofusion}, OmniControl~\cite{xie2023omnicontrol}, and CondMDI~\cite{cohan2024flexible}), we adapt them by replacing their text encoders with a Vision Transformer (ViT)-based global scene encoder to incorporate scene conditions.
For diffusion-based motion synthesis methods (MDM~\cite{tevet2023human}, StableMoFusion~\cite{huang2024stablemofusion}, and SceneDiffuser~\cite{SceneDiffuser}), we modify their inference process to support motion in-betweening by imputing joint positions at every diffusion step. Additionally, we adapt their motion representations to incorporate a global root representation, enabling keyframe-based in-betweening via imputation sampling.
Across all baselines, we use only static keyframe poses—such as joint position information—to generate intermediate motions.

\section{Additional Analysis}
\label{sec:add_details}

\subsection{Robustness to Varying Keyframe Selection Strategy} 

Our motion in-betweening module experiences random keyframes with mask $\mathbf{m}$ during training, it maintains strong performance with arbitrary keyframes $\mathbf{m}^*$ at the inference.
We show that our method consistently achieves robust results with keyframes chosen at arbitrary indices, even in noisy conditions.
Table~\ref{tab:experiments_diverse_keyframes} demonstrates the robustness of our method across different keyframe selection strategies, showing its ability to handle noise effectively.
\input{tab/mib_diverse_keyframe}

\subsection{Time Cost}

We report the inference time comparison with baselines in Table~\ref{table:inference_speed} for obtaining SMPL parameters. For realistic character animation, acquiring actual motion parameters is essential. Our method directly predicts these parameters, whereas baselines require a post-processing with an additional optimization-based fitting process from predicted joint positions. This offers a faster pipeline for obtaining actual motion compared to baselines.

\input{tab/infertime}

\subsection{Ablation on Hyper Parameters}
\label{sec:suppl_ablation_hyperparameters}

We evaluate different configurations of global scene dimensions, the number of BPS points, and body shape conditioning within a sparse keyframe interval setup ($r=60$) to validate our hyperparameter choices in Table~\ref{tab:experiments_hyper}.

We design body shape encoding, $\mathbf{b}$, that includes key joint-to-joint distances from T-pose: $[\texttt{\small root}, \texttt{\small head}]$, $[\texttt{\small left\_shoulder}, \texttt{\small right\_shoulder}]$, $[\texttt{\small shoulder}, \texttt{\small wrist}]$, $[\texttt{\small left\_pelvis}, \texttt{\small right\_pelvis}]$, and $[\texttt{\small pelvis}, \texttt{\small feet}]$. Two thickness values: distances between the frontmost and rearmost vertices of the chest region and the hip region. These measurements provide a body shape abstraction $\mathbf{b}$ as a compact shape representation in a continuous domain.
Furthermore, our main experiments are conducted on diverse body shapes, including five samples from the TRUMANS dataset and real-world shapes from GIMO and Video2Animation. Although the design of the body shape encoding is not our primary contribution, it significantly enhances in-betweening accuracy and reduces penetration artifacts.

\input{tab/mib_ablation_hyper}

\input{sec/X_suppl_video2animation}

%% file: tab/hyperparameter.tex
\begin{table}[h]
    \centering
    \resizebox{0.85\columnwidth}{!}{
    \begin{tabular}{cc}
        \toprule
        \multicolumn{1}{p{5.1cm}}{\centering Hyperparameter} & \multicolumn{1}{p{2.1cm}}{\centering Value}
        \\
        \midrule
        Batch size & 256 \\
        Learning rate & 1e-4 \\
        Optimizer & Adam W\\
        Weight decay &  1e-2\\
        Channels dim &  256 \\
        Channel multipliers &  $[2,2,2,2]$\\
        Variance scheduler & Cosine~\cite{nichol2021improveddenoisingdiffusionprobabilistic}\\
        Diffusion steps &  1000\\
        Diffusion variance & $\tilde{\beta} = \frac{1-\alpha_{t-1}}{1-\alpha_t}\beta_t$\\
        EMA weight ($\beta$)  & 0.9999\\
        Guidance weight ($\mathbf{w}$)  & 2.5\\
        \bottomrule
    \end{tabular}
    }
    \caption{Hyperparameters of the Model}
    \vspace{-1.0em}
    \label{table:hyperparameters}
\end{table}

%% file: tab/mib_diverse_keyframe.tex
\begin{table}[h!]
    \centering
\resizebox{0.99\linewidth}{!}{
\begin{tabular}{lccc}
\toprule
\multicolumn{1}{p{2.8cm}}{\centering Keyframe Selection} & \multicolumn{1}{p{0.9cm}}{\centering FID $\downarrow$} & \multicolumn{1}{p{2.0cm}}{\centering Jerk $(m/s^3)$ $\downarrow$} & \multicolumn{1}{p{2.25cm}}{\centering MJPE All ($m$) $\downarrow$} \\
\midrule
Uniform $(r=1)$ & 0.122 & 0.197 & 0.0117   \\
Uniform $(r=3)$ & 0.118 & 0.198 & 0.0129    \\
Uniform $(r=15)$ & 0.125 & 0.196 & 0.0153   \\
Uniform $(r=60)$  & 0.123 & 0.198 & 0.0233 \\
\midrule
Random $(p=0.2)$ & 0.124 & 0.199 & 0.0138 \\
Random $(p=0.5)$ & 0.123 & 0.199 & 0.0124 \\
 \bottomrule
\end{tabular}
}
\caption{Quantitative evaluation of diverse keyframe selection strategies on noisy TRUMANS test set with a fixed noise level $l=1$. We select keyframes using different strategies, such as at a uniform interval $r$ or with a random probability $p$, including start and end frames. Our method shows robustness performance from highly sparse to dense keyframes, regardless of keyframe density or selection.
}
\label{tab:experiments_diverse_keyframes}
\vspace{-1.5em}
\end{table}

%% file: tab/infertime.tex
\begin{table}[h!]
    \centering
    \resizebox{\columnwidth}{!}{
    \begin{tabular}{ccccc}
        \toprule
        Method & MDM~\cite{tevet2023human} & OmniControl~\cite{xie2023omnicontrol} & CondMDI~\cite{cohan2024flexible} & Ours
        \\
        \midrule
        Time (s) & 119.4 $\pm$ 2.1  & 283.7 $\pm$ 3.8 & 162.4 $\pm$ 3.5 & 39.6 $\pm$ 0.8 \\
        \bottomrule
    \end{tabular}
    }
    \caption{Time required to obtain actual parameters for motion.}
    \label{table:inference_speed}
    \vspace{-0.86em}
\end{table}

%% file: tab/mib_ablation_hyper.tex
\begin{table}[ht!]
    \centering
\resizebox{1.0\linewidth}{!}{
\begin{tabular}{lccccc}
\toprule
\multirow{2}{*}{\centering Configurations} & \multirow{2}{*}{\centering FID~$\downarrow$} & \multicolumn{1}{p{1.5cm}}{\centering Jerk ($m/s^3$) $\downarrow$}  & \multicolumn{1}{p{1.6cm}}{\centering MJPE All ($m$) $\downarrow$} & \multicolumn{1}{p{2.3cm}}{\centering Collision Frame Ratio  $\downarrow$} & \multicolumn{1}{p{1.8cm}}{\centering Pene. Max ($m$) $\downarrow$} \\
\midrule
Scene 96x48x96 & 0.130 & 0.201 &0.027 & 0.117 & 0.046 \\
BPS 256         & 0.124 & 0.196 & 0.025 & 0.114 & 0.045    \\
w/o Body Shape & \textbf{0.122} & \textbf{0.193} &  0.038 & 0.121 & 0.047  \\
Ours           & 0.123 & 0.194 &  \textbf{0.023}   & \textbf{0.113} & \textbf{0.043}   \\
\bottomrule
\end{tabular}
}
\caption{Ablation study on our hyperparmeters setting.}
\label{tab:experiments_hyper}
\vspace{-1.2em}
\end{table}

%% file: sec/X_suppl_video2animation.tex
\section{Video-based Human-Scene Interaction Reconstruction}
\label{sec:suppl_video2animation}

In this section, we present a Human-Scene Interaction Reconstruction pipeline, where our SceneMI module plays a core component.
The goal is to reconstruct realistic, physically plausible human animations and scene geometry from monocular RGB video sequences that capture both scene and human movements.

The pipeline comprises two primary stages: the \textit{initial stage} and the \textit{refinement stage}.
In the \textit{initial stage}, we extract a rough estimate of both human motion and scene geometry in a metric scale.
In the \textit{refinement stage}, we enhance the physical plausibility and naturalness of the motion using the reconstructed scene geometry and our SceneMI module.
The following sections detail the challenges and methodologies for each stage.

\subsection{Initial Stage}

Our framework takes as input an RGB video sequence of $M$ frames with 30 FPS, denoted as $\{I_i\}_{i=1}^M$.

\paragraph{Camera Parameter Estimation}
From the first frame of the video sequence, we estimate intrinsic camera parameters using~\cite{jin2023perspective}. These parameters are crucial for positioning 3D human meshes or back-projecting depth estimation results in subsequent steps.

\paragraph{Human Mesh Recovery (HMR)}

We utilize 4D Humans~\cite{goel2023humans} to obtain human mesh parameters for each frame. The obtained parameters are used to construct SMPL model-based human meshes, denoted as $\{X_i\}_{i=1}^M$. These meshes are then placed in 3D space using the previously estimated camera parameters and root translations. Since the SMPL model is defined in metric scale~\cite{SMPL:2015}, this process provides an initial metric-scale geometry reference.

\paragraph{Metric-Scale Depth Estimation with HMR}

To recover the complete 3D scene geometry, we employ a pre-trained depth estimation network~\cite{depthanything} to produce initial depth maps $D_{\text{init},i}$ for each frame. These depth maps, while precisely capturing relative depth relationships, lack accurate metric-scale representation. To resolve this, we estimate a global scale $s$ and offset factor $o$ that transform the $D_{\text{init},i}$ into metric-scale:
$$
D_i = s \cdot D_{\text{init},i} + o, \quad \forall i = 1, 2, \dots, M
$$

To determine the optimal transformation parameters $s$ and $o$, we leverage the metric-scale human meshes ($X_i$) obtained in the previous stage as geometric references. For each frame $i$, we sample the visible vertices from the human mesh in camera space, denoted as $V(X_i)$. We also backproject the transformed depth map $D_i$ into 3D space, selecting only the region corresponding to human segmentation in image $I_i$, to obtain point clouds denoted as $P_X(D_i)$. The alignment between these point sets is achieved by minimizing the chamfer distance between two pointsets:
$$
\mathcal{L} = \sum_{i=1}^M  d(V(X_i), P_X(D_i))
$$
where $d$ represents the Chamfer distance~\cite{ravi2020pytorch3d} between two point sets.
Optimization ensures that the transformed depth maps align with the metric-scale geometry of human models.

However, depth estimation results are often uncertain, particularly at object boundaries. To address this, we estimate the uncertainty of depth values and retain only reliable information. We apply color jittering transformations (hue transformations)~\cite{torchvision2016} to the input image and obtain multiple depth values for each pixel. We calculate uncertainty following~\cite{10328050} and only valid depth values are preserved for subsequent steps.

\paragraph{Reconstruct Individual Objects}
To reconstruct the 3D scene, we adopt a strategy that restores individual objects from the video as 3D meshes $M_j$ and places them accurately within the 3D space.
Our process begins by obtaining instance segmentation~\cite{cheng2021mask2former} results from the provided video frames. However, due to occlusions caused by foreground objects or human movement, these initial segmentation results are often incomplete or imprecise. We address this limitation by employing an image completion algorithm~\cite{ozguroglu2024pix2gestalt} to refine the segmentation and generate a more complete image for each object. Given these refined segmentation results, we then apply an Image-to-3D reconstruction method~\cite{xu2024instantmesh, yang2024tencent} to obtain initial 3D object meshes $M_j$ with textures for each instance.

\paragraph{Object Scale and Pose Refinement}

Individually reconstructed objects $M_j$ exhibit inaccuracies in scale and pose.
Empirically, we observe that reconstructed objects align well with the gravity, but require refinement in translations $t_j$, rotations $r_j$, and scales $s_j$.
We address these issues by optimizing it using metric-scale depth maps $D$.

For each object mesh $M_j$, we sample visible surface points in camera space, denoted as $V(M_j)$. Then, these points are transformed using a learnable variable $t_j$, $r_j$, and $s_j$.
We also extract corresponding points from the metric-scale depth map $D$ using the object's segmentation mask, denoted as $P_j(D)$. After initializing the object's translation $t_j$ using the centroid of $P_j(D)$, we optimize the object's scale $s_j$, rotations $r_j$, and translation $t_j$ by minimizing:
$$
\mathcal{L} = d(V(M_j), P_j(D))
$$
where $d$ represents the Chamfer distance between two point sets.

\begin{figure}[h]
\centering
\includegraphics[width=1.0\linewidth]{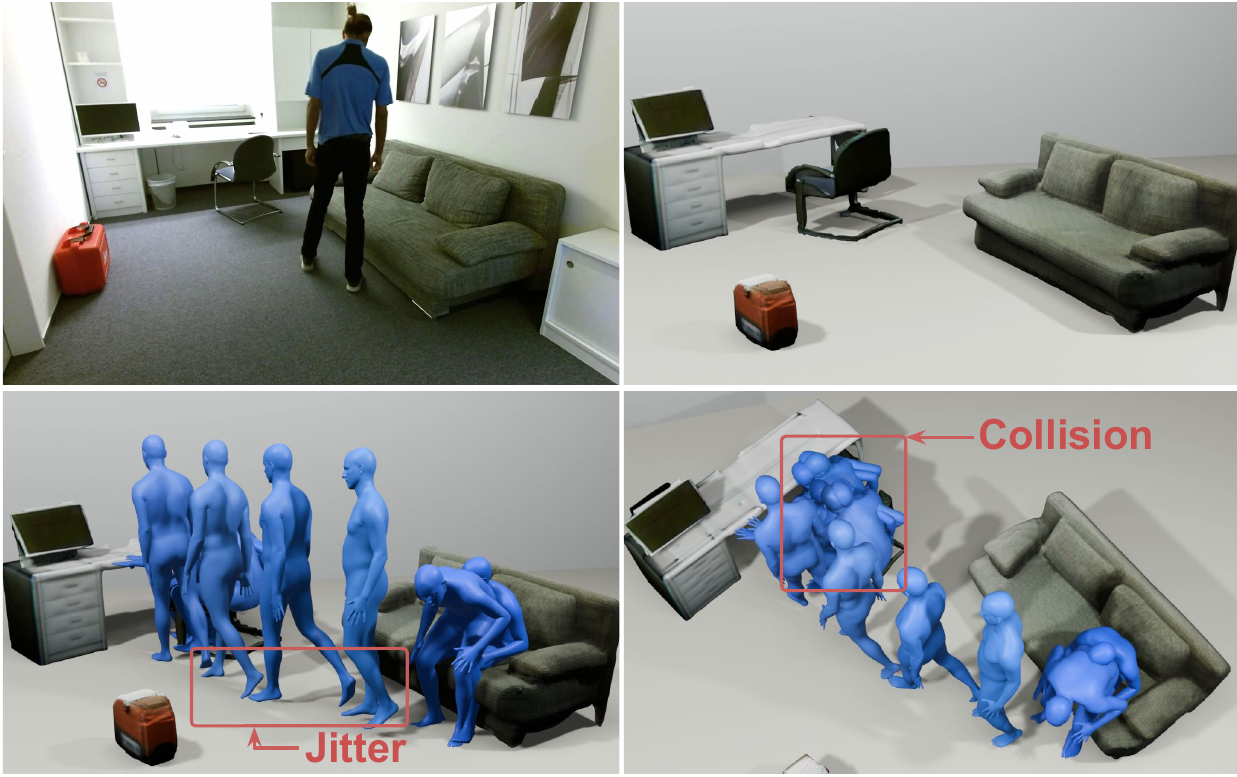}
\caption{Results from the \textit{initial stage} of Video2Animation. Starting from the input video~\cite{PROX} (top left), we reconstruct the scene geometry (top right) and the corresponding human motion (bottom) in metric scale.}
\label{fig:video2animation_init}
\vspace{-0.3em}
\end{figure}

\subsection{Refinement Stage}

Following the initial stage of motion and scene geometry reconstruction in a metric scale, several challenges remain in motion estimation, including potential scene collisions, motion jittering, and inconsistencies inherent to image-based motion extraction algorithms, as shown in Figure~\ref{fig:video2animation_init}. We address these issues by leveraging a 3D motion prior by applying our SceneMI module.

\paragraph{Keyframe Optimization}

We optimize keyframes at regular 5-frame intervals, concentrating on root translation $\gamma$ where motion estimation errors predominantly occur. The optimization leverages five complementary loss functions:

\textit{Regularization Loss} constraints large deviations from the initial guess, ensuring optimization stability.
\textit{Contact Loss} estimates contact vertices~\cite{zhang2020phosa} from human meshes $X_i$, encouraging precise alignment with scene geometry while penalizing non-contact vertex penetrations.
\textit{Temporal Smoothing Loss} minimizes consecutive root translation differences, encouraging smooth transitions between frames.
\textit{Depth Matching Loss} aligns visible human mesh points with metric-scale depth estimations using Chamfer distance minimization.
\vspace{-0.8em}

\paragraph{Applying SceneMI}

\begin{figure}[h]
\centering
\includegraphics[width=1.0\linewidth]{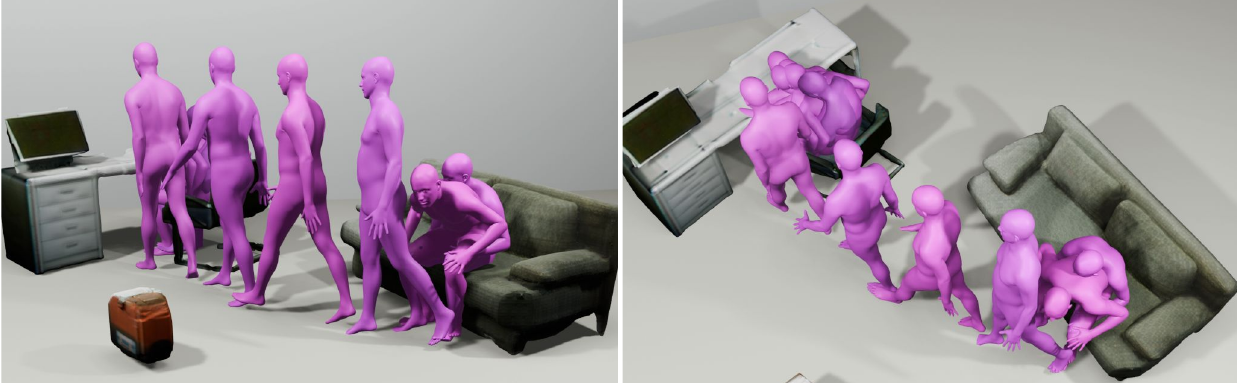}
\caption{The final results from the Video2Animation pipeline demonstrate the reconstruction of 3D human-scene animation from monocular video inputs. By incorporating SceneMI with the obtained scene information and optimized keyframes, we reconstruct natural and physically plausible motions. For additional results, please refer to the supplementary video.
}
\label{fig:video2animation_after}
\vspace{-1em}
\end{figure}

Following keyframe optimization, we progressively refine overall motion sequences using SceneMI. We sample one keyframe from every three optimized keyframes, corresponding to a 15-frame interval in the original video. By leveraging scene geometry and the poses derived from keyframes, we reconstruct the final animation that integrates geometric constraints, enhancing both realism and physical plausibility, as shown in Figure~\ref{fig:video2animation_after}.

As SceneMI limits motion sequence synthesis to length $N=121$, we employ an autoregressive strategy to synthesize continuous and natural human motion across extended sequences. For keyframes representing arbitrary motion lengths, we divide sequences into $N$-length segments with $v$ frame overlaps, where $v = 60$.
We iteratively synthesize motion by using the final $v$ frames of a prior episode as initial keyframes for the subsequent segment. After synthesizing the first motion sequence, we utilize its last $v$ frames as keyframes for the start of the subsequent segment. For the remaining $N-v$ frames, motion is synthesized based on the corresponding keyframes from the current segment.

This progressive approach enables motion synthesis across long sequences, overcoming SceneMI's length constraints while maintaining scene awareness and motion consistency.
This autoregressive approach allows applicability to real-world videos with arbitrary-length inputs.

\section{Additional Results}
\label{sec:add_results}

\subsection{Evaluation across Multiple Seeds} 

Since our model is generative, we repeat our major experiments in Table 1 and Table 3 in the main paper across 20 different random seeds. We report their mean value, with 95\% statistical confidence interval in Table~\ref{tab:experiments_multiple_seed}. The observed variance is marginal, demonstrating the stability of our method.

\begin{table}[h!]
    \centering
\resizebox{1.0\linewidth}{!}{
\begin{tabular}{lllllll}
\toprule
\multirow{2}{*}{\centering  Configurations} & \multirow{2}{*}{\centering FID~$\downarrow$} &  \multicolumn{1}{p{1.6cm}}{\centering Jerk ($m/s^3$) $\downarrow$} & \multirow{2}{*}{\centering MJPE ($m$) $\downarrow$} & \multirow{2}{*}{\centering Collision Ratio $\downarrow$} \\

\midrule
Tab.1 (ours) & 0.123 & 0.194  & 0.023 & 0.113 \\

\quad + 20 runs & \et{0.123}{0.001} & \et{0.193}{0.002}  & \et{0.023}{0.001} & \et{0.114}{0.002}  \\
\midrule
Tab.3 (ours) & 0.118 & 0.198  & 0.012 & 0.108  \\
\quad + 20 runs & \et{0.118}{0.001} & \et{0.198}{0.003}  & \et{0.012}{0.001} & \et{0.109}{0.002}  \\
\bottomrule
\end{tabular}
}
\caption{Evaluation across multiple random seeds. We report the mean and 95\% confidence intervals for key metrics over 20 runs.}
\label{tab:experiments_multiple_seed}
\end{table}
\vspace{-1em}

\subsection{Integration with Frame-Based HSI} 
\begin{figure}[h!]
\centering
\vspace{-0.5em}
\includegraphics[width=0.92\linewidth]{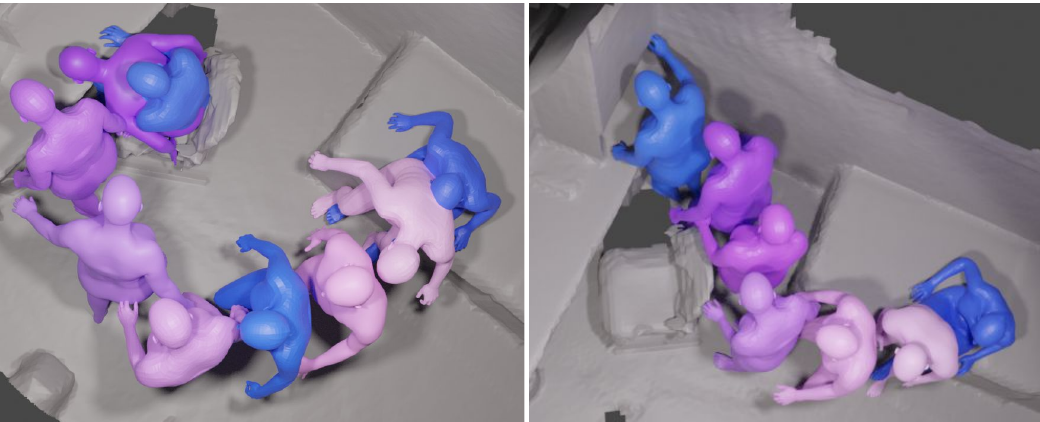}
\vspace{-0.5em}
\caption{Integration with semantically generated keyframes. Our model produces plausible motions from sparse, semantic keyframes.}
\label{fig:unseen}
\end{figure}

To further explore the applicability of our method, we integrate our module with a semantic keyframe generation approach. Specifically, we generated multiple sparse keyframes (colored in \textcolor{blue}{blue}) using COINS~\cite{Zhao2022coins} to provide semantic cues in various scenes, then applied our model to synthesize the complete motion sequence. The Figure~\ref{fig:unseen} show our method generates coherent and plausible motions despite the semantic sparsity of the input keyframes.

\vspace{-1em}

\subsection{Long-Term Keyframe Interval}  
\begin{figure}[h!]
\centering
\includegraphics[width=0.92\linewidth]{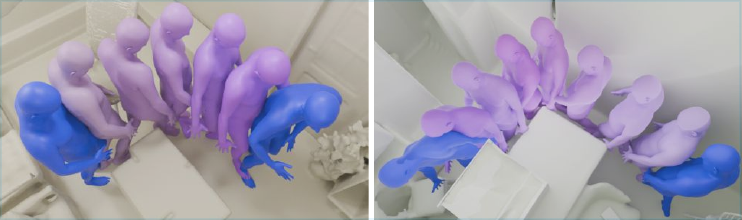}
\vspace{-0.5em}
\caption{Result with a long-term keyframe interval. The model synthesizes long-horizon motion while avoiding large obstacles.}
\label{fig:long_interval}
\end{figure}

We also additionally provide an example with a 4-second keyframe interval, where only the start and end frames are given. As shown in Figure~\ref{fig:long_interval}, the synthesized motion successfully navigates complex scenes with large obstacles, demonstrating our model successfully handles a long motion sequence that navigates around large obstacles.

\vspace{-0.9em}

\subsection{Discussion of Failure Cases} 
\begin{figure}[h!]
\centering
\includegraphics[width=0.95\linewidth]{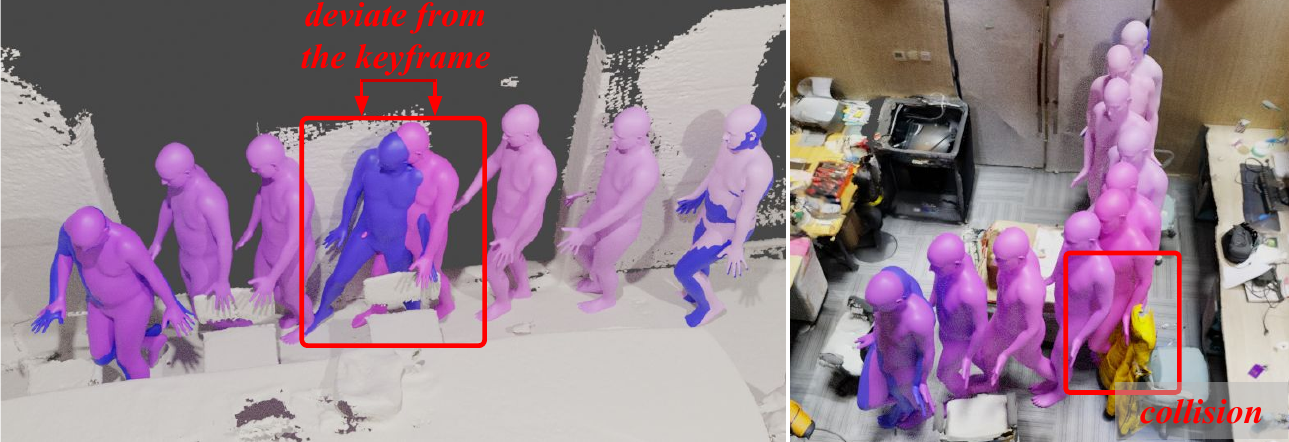}
\caption{Failure cases. (Left) Unseen interaction pattern (e.g., squeezing through narrow space). (Right) Real-world scene with noisy or complex geometry.}
\label{fig:failure}
\end{figure}

While our method generalizes well to unseen configurations and scene geometries beyond the training data, we acknowledge certain failure cases. First, failure can occur in rare human-scene interaction patterns such as “squeezing through a narrow passage,” where the required motion rarely observed in training. Second, performance degrades in real-world scenes with highly complex or noisy geometric reconstructions, where subtle spatial constraints may not be fully captured by scene encoding. Figure~\ref{fig:failure} illustrates representative failure cases.

